\documentclass[11pt,a4paper]{article}
\usepackage[hyperref]{acl2021}
\usepackage{times}
\usepackage{latexsym}

\aclfinalcopy 

\usepackage{microtype}
\usepackage{amssymb}
\usepackage{amsmath}
\DeclareMathOperator*{\argmax}{argmax}

\usepackage{cancel}
\usepackage{soul}
\usepackage{url}
\usepackage{graphicx}
\usepackage{amsthm}
\usepackage{booktabs}
\usepackage{latexsym}
\usepackage{comment}
\usepackage{color}
\usepackage{xcolor}
\usepackage{commath}
\usepackage{multirow}
\usepackage{enumitem}

\newcommand{\RNum}[1]{\uppercase\expandafter{\romannumeral #1\relax}}

\usepackage{helvet}  
\usepackage{courier}  
\usepackage{tabularx}  
\usepackage{color}
\usepackage{CJKutf8}
\usepackage{makecell}
\usepackage{bigstrut}
\usepackage{tabu}
\usepackage{multirow}
\usepackage[normalem]{ulem}
\usepackage{subfigure}

\newcommand{\ie}{\textit{i}.\textit{e}.}
\newcommand{\eg}{\textit{e}.\textit{g}.}

\title{Robustifying Multi-hop Question Answering through \\ Pseudo-Evidentiality Training }

\author{Kyungjae Lee$^1$ \quad  Seung-won Hwang$^2$\thanks{~~correspond to seungwonh@snu.ac.kr} \quad  Sang-eun Han$^1$ \quad  Dohyeon Lee$^1$  \\ $^1$Yonsei University ~~ \quad $^2$Seoul National University}

\begin{document}
\maketitle

\begin{abstract}

This paper studies the bias problem of multi-hop question answering models, of answering correctly without correct reasoning. One way to robustify these models is by supervising to not only answer right, but also  with right reasoning chains. An existing direction is to annotate reasoning chains to train models, requiring expensive additional annotations. In contrast, we propose a new approach to learn evidentiality, deciding whether the answer prediction is supported by correct evidences, without such annotations. Instead, we compare counterfactual changes in answer confidence with and without evidence sentences, to generate ``pseudo-evidentiality" annotations. We validate our proposed model on an original set and challenge set in HotpotQA, showing that our method is accurate and robust in multi-hop reasoning.

\end{abstract}

\section{Introduction}

Multi-hop Question Answering (QA) is a task of answering complex questions by connecting information from several texts.
Since the information is spread over multiple facts, this task requires to capture
multiple relevant facts (which we refer as evidences) and infer an answer based on all these evidences.

However, previous works \cite{min2019compositional,chen2019understanding,trivedi2020multihop} observe ``disconnected reasoning" in some correct answers.
It happens when models can exploit specific types of artifacts (\eg, entity type),
to leverage them as \textbf{reasoning shortcuts} to guess the correct answer.
For example, assume that a given question is: ``which country got independence when World War \RNum{2} ended?'' and a passage is: ``Korea got independence in 1945".
Although information (``World War \RNum{2} ended in 1945") is insufficient, QA models predict ``Korea",
simply because its answer type is country (or, using shortcut).

\begin{figure}[t]
	\centering
	\includegraphics[width=68mm]{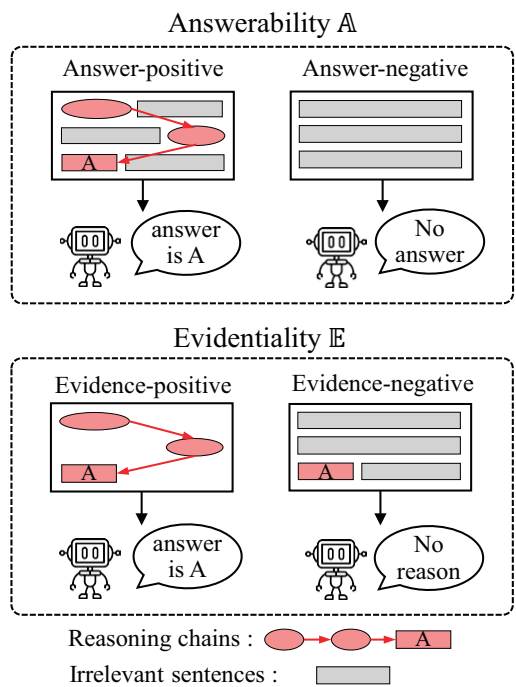}
	\caption{Overview of our proposed supervision: using Answerability and Evidentiality}
	\label{figure1}
\end{figure}

To address the problem of reasoning shortcuts,
we propose to supervise ``\textbf{evidentiality}"  -- deciding whether a model answer is supported by correct evidences (see Figure~\ref{figure1}).
This is related to the problem that most of the early reader models
for QA failed to predict whether questions are not answerable. 
Lack of answerability training led models to provide a wrong answer with high confidence, when they had to answer ``unanswerable".
Similarly, we aim to train for models to recognize whether their answer is ``unsupported" by evidences, as well.
In our work, along with the answerability,  
we train the QA model to identify the existence of evidences by using passages of two types: (1) \textbf{Evidence-positive} and (2) \textbf{Evidence-negative} set.
While the former has both answer and evidence, 
the latter does not have evidence supporting the answer, such that we can detect models taking shortcuts.

Our first research question is: how do we \textbf{acquire} evidence-positive
and negative examples for training without annotations?
For evidence-positive set, the closest existing approach \cite{niu2020self} is to consider attention scores, which can be considered as pseudo-annotation for evidence-positive set.
In other word, sentence $S$ with high attention scores, often used as an ``interpretation" of whether $S$ is causal for
model prediction, can be selected to build evidence-positive set.
However, follow-up works \cite{serrano2019attention,jain2019attention} argued that attention is limited as an explanation, because causality cannot be measured, without observing model behaviors in a counterfactual case of the same passage without $S$.
In addition, sentence causality should be aggregated to measure group causality of multiple evidences for multi-hop reasoning.
To annotate group causality as ``pseudo-evidentiality", we propose  \textit{Interpreter}  module, which
removes and aggregates evidences into a group, to compare predictions in observational and counterfactual cases.

As a second research question, we ask how to \textbf{learn} from evidence-positive and evidence-negative set.
To this end, we identify two objectives: (O1) QA model should not be overconfident in evidence-negative set, while (O2) confident in evidence-positive.
A naive approach to pursue the former is to lower the model confidence on evidence-negative set via regularization.
However, such regularization can cause violating (O2) due to correlation between confidence distributions for evidence-positive and negative set.
Our solution is to selectively regularize, by purposedly training a biased model violating (O1), 
and decorrelate the target model from the biased model.

For experiments, we demonstrate the impact of our approach on HotpotQA dataset.
Our empirical results show that our model can improve QA performance through pseudo-evidentiality, outperforming other baselines.
In addition, our proposed approach can orthogonally combine with another SOTA model for additional performance gains.

\section{Related Work}

Since multi-hop reasoning tasks, such as 
HotpotQA, are released, many approaches for the task have been proposed.
These approaches can be categorized by strategies used, such as graph-based networks~\cite{qiu2019dynamically,fang2019hierarchical}, external knowledge retrieval~\cite{asai2019learning}, and supporting fact selection~\cite{nie2019revealing,groeneveld2020simple}.

Our focus is to identify and alleviate reasoning shortcuts in multi-hop QA, without evidence annotations.
Models taking shortcuts were widely observed from various tasks, such as object detection~\cite{singh2020don}, NLI~\cite{tu2020empirical}, and also for our target task of multi-hop QA~\cite{min2019compositional,chen2019understanding,trivedi2020multihop}, 
where models learn simple heuristic rules, answering correctly but without proper reasoning.

To mitigate the effect of shortcuts,
adversarial examples~\cite{jiang2019avoiding}
can be generated, or alternatively, models can be robustifed~\cite{trivedi2020multihop} 
with additional supervision for paragraph-level ``sufficiency" -- to identify whether a pair of two paragraphs are sufficient for right reasoning or not, which reduces shortcuts on a single paragraph.
While the binary classification for paragraph-sufficiency is relatively easy (96.7 F1 in \citet{trivedi2020multihop}), 
our target of capturing a finer-grained sentence-evidentiality is more challenging. 
Existing QA model~\cite{nie2019revealing,groeneveld2020simple} treats this as a supervised task, based on sentence-level human annotation.
In contrast, ours requires no annotation and focuses on avoiding reasoning shortcuts using evidentiality, which was not the purpose of evidence selection in the existing model.

\section{Proposed Approach}

In this section, to prevent reasoning shortcuts, we introduce a new approach for data acquiring and learning.
We describe this task (Section 3.1) and address two research questions, of generating labels for supervision (Section 3.2) and learning (Section 3.3), respectively.

\subsection{Task Description}

Our task definition follows \textit{distractor} setting,
between \textit{distractor} and \textit{full-wiki} in HotpotQA dataset~\cite{yang2018hotpotqa}, which consists of 112k questions requiring the understanding of corresponding passages to answer correctly.
Each question has a candidate set of 10 paragraphs (of which two are positive paragraphs $\mathcal{P}^+$ and eight are negative $\mathcal{P}^-$), where the supporting facts for reasoning are scattered in two positive paragraphs.
Then, given a question $\mathcal{Q}$, the objective of this task is to aggregate relevant facts from the candidate set and estimate a consecutive answer span $\mathcal{A}$.
For task evaluation, the estimated answer span is compared with the ground truth answer span in terms of F1 score at word-level.

\subsection{Generating Examples for Training Answerability and Evidentiality}

\subsubsection*{Answerability for Multi-hop Reasoning}

For answerability training in single-hop QA, 
datasets such as SQuAD 2.0~\cite{rajpurkar2018know} provide labels of answerability, so that models can be trained not to be overconfident on unanswerable text.

Similarly, we build triples of question $\mathcal{Q}$, answer $\mathcal{A}$, and passage $\mathcal{D}$, to be labeled for answerability.
HotpotQA dataset pairs $\mathcal{Q}$ with 10 paragraphs,
where evidences can be scattered to two paragraphs.
Based on such characteristic, concatenating two positive paragraphs is guaranteed to be answerable/evidential  and concatenating two negative paragraphs (with neither evidence nor answer) is guaranteed to be unanswerable.
We define a set of answerable triplets ($\mathcal{Q,A,D}$) 
as \textbf{answer-positive} set $\mathbb{A}^+$, 
and an unanswerable set
as \textbf{answer-negative} set $\mathbb{A}^-$.
From the labels, we train a transformer-based model to classify the answerability (the detail will be discussed in the next section).

However, answerability cannot supervise whether the given passage has 
all of these relevant evidences for reasoning.
This causes a lack of generalization ability, especially on examples with an answer but no evidence.

\subsubsection*{Evidentiality for Multi-hop Reasoning}
While learning the answerability, we aim to capture the existence of reasoning chains in the given passage.
To supervise the existence of evidences, we construct examples:
\textbf{evidence-positive} and \textbf{evidence-negative} set, as shown in Figure~\ref{figure1}.

Specifically, let $\mathrm{\textit{E}}_*$ be the ground truth of evidences to infer $\mathcal{A}$, and $\mathcal{S}_*$ be a sentence containing an answer $\mathcal{A}$, corresponding to $\mathcal{Q}$.
Given $\mathcal{Q}$ and $\mathcal{A}$, 
expected labels $\mathcal{V}_E$ of evidentiality, indicating whether the evidences for answering are sufficient in the passage, are as follow:
\begin{equation}
    \label{evidentiality}
    \begin{aligned}
        & \mathcal{V}_E (\mathcal{Q,A,D}) \models True  & \Leftrightarrow ~~  \mathrm{\textit{E}}_* = \mathcal{D}, ~ \mathcal{A} \subset \mathcal{D}  \\
        & \mathcal{V}_E (\mathcal{Q,A,D}) \models False  & \Leftrightarrow ~~  \mathrm{\textit{E}}_* \not\subset \mathcal{D}, ~ \mathcal{A} \subset \mathcal{D} \\
    \end{aligned}
\end{equation}
We define a set of passages satisfying $\mathcal{V}_E \models True$ as  \textbf{evidence-positive} set $\mathbb{E}^+$, and a set satisfying $\mathcal{V}_E \models False$ as \textbf{evidence-negative} set $\mathbb{E}^-$.

Since we do not use human-annotations, we aim to generate ``pseudo-evidentiality" annotation.
First, for \textbf{evidence-negative} set, 
we modify answer sentence $\mathcal{S}_*$ and unanswerable passages, and 
generate examples with the three following types:
\begin{itemize}[leftmargin=0.6cm]
    \item 1) Answer Sentence Only: we remove all sentences in answerable passage except $\mathcal{S}_*$, such that the input passage $\mathcal{D}$ becomes $\mathcal{S}_*$, which contains a correct answer but no other evidences. That is, $\mathcal{V}_E (\mathcal{Q,A,S^*}) \models  False $.
    \item 2) Answer Sentence + Irrelevant Facts: we use irrelevant facts with answers as context, by concatenating $\mathcal{S}_*$ and unanswerable $\mathcal{D}$. That is, $\mathcal{V}_E (\mathcal{Q,A,(S^*;D)}) \models  False $, where $\mathcal{D} \in \mathcal{P}^-$.
    \item 3) Partial Evidence + Irrelevant Facts: we use partially-relevant and irrelevant facts as context, by concatenating $\mathcal{D}_1 \in \mathcal{P}^+$ and $\mathcal{D}_2 \in \mathcal{P}^-$. That is, $\mathcal{V}_E (\mathcal{Q,A,} (\mathcal{D}_1;\mathcal{D}_2)) \models  False $.
\end{itemize}
These \textbf{evidence-negative} examples do not have all relevant evidences, thus if a model predicts the correct answer on such examples, it means that the model learned reasoning shortcuts.

Second, building an \textbf{evidence-positive} set is more challenging, because it is difficult to capture multiple relevant facts, with neither annotations $E_*$ nor supervision.
Our distinction is obtaining the above annotation from model itself, by interpreting the internal mechanism of models.
On a trained model, we aim to find influential sentences in predicting correct answer $\mathcal{A}$, among sentences in an answerable passage.
Then, we consider them as a pseudo evidence-positive set.
Since such pseudo labels relies on the trained model which is not perfect, 
100\% recall of $\mathcal{V}_E (\mathcal{Q,A,D}) \models True$ in Eq. (\ref{evidentiality}) is not guaranteed, though we observe 87\% empirical recall (Table \ref{statistic}).

Section 1 discusses how interpretation, such as 
attention scores~\cite{niu2020self}, can be pseudo-evidentiality.
For QA tasks, an existing approach \cite{perez2019finding} uses answer confidence for finding pseudo-evidences, as we discuss below:

(A) Accumulative interpreter:
to consider multiple sentences as evidences, the existing approach \cite{perez2019finding} iteratively inserts  sentence $\mathcal{S}_i$ into set $\mathrm{\textit{E}}^{t-1}$,  with a highest probability at $t$-th iteration, as follows:
\begin{equation}
    \begin{aligned}
       & \Delta P_{\mathcal{S}_i} = ~ P(\mathcal{A}|\mathcal{Q}, \mathcal{S}_i \cup \mathrm{\textit{E}}^{t-1}) - P(\mathcal{A}|\mathcal{Q},\mathrm{\textit{E}}^{t-1} ) \\
       & ~~~~~ \hat{\mathrm{\textit{E}}}^t = \argmax_{\mathcal{S}_i} \Delta P_{\mathcal{S}_i},  ~~~~~ \mathrm{\textit{E}}^t =  \hat{\mathrm{\textit{E}}}^t \cup \mathrm{\textit{E}}^{t-1}  \\
    \end{aligned}
    \label{iterative}
\end{equation}
where $\mathrm{\textit{E}}^0$ starts with the sentence $\mathcal{S}_*$ containing answer $\mathcal{A}$, which is minimal context for our task.
This method can consider multiple sentences as evidence by inserting iteratively into a set, but cannot consider the effect of \textbf{erasing} sentences from reasoning chain.

(B) Our proposed \textit{Interpreter}: 
to enhance the interpretability, we  consider both \textbf{erasing} and \textbf{inserting} each sentence, in contrast to accumulative interpreter considering only the latter.
Intuitively, erasing evidence would change the prediction significantly,
if such evidence is causally salient, 
which we compute as follows:
\begin{equation}
\label{our_erasure}
  \Delta P_{\mathcal{S}_i} = P(\mathcal{A}|\mathcal{Q,D}) -  P(\mathcal{A}|\mathcal{Q}, (\mathcal{D} \backslash \mathcal{S}_i))
\end{equation}
where $(\mathcal{D} \backslash \mathcal{S}_i)$ is a passage out of sentence $\mathcal{S}_i$. 
We hypothesize that breaking reasoning chain, by erasing $\mathcal{S}_i$,
should significantly decrease $P(A|\cdot)$.
In other words, 
$\mathcal{S}_i$ with higher $\Delta P_{\mathcal{S}_i}$ is salient.
Combining the two saliency scores in Eq. (\ref{iterative}),(\ref{our_erasure}), our final saliency is as follows:
\begin{equation}
    \begin{aligned}
        & ~~ \Delta P_{\mathcal{S}_i} =  ~ P(\mathcal{A}|\mathcal{Q}, \mathcal{S}_i \cup \mathrm{\textit{E}}^{t-1}) - 
        \cancel{P(\mathcal{A}|\mathcal{Q},\mathrm{\textit{E}}^{t-1} )} \\
        & + \cancel{P(\mathcal{A}|\mathcal{Q,D})} - P(\mathcal{A}|\mathcal{Q}, (\mathcal{D} \backslash (\mathcal{S}_i \cup \mathrm{\textit{E}}^{t-1})) ) \\
    \end{aligned}
    \label{interpreter}
\end{equation}
where the constant values can be omitted in $\argmax$.
At each iteration, the sentence that maximize $\Delta P_{\mathcal{S}_i}$ is selected, as done in Eq. (\ref{iterative}).
This promotes selection that increases confidence $P(\mathcal{A}|\cdot)$ on important sentences, and decreases confidence on unimportant sentences.
We stop the iterations if $\Delta P_{\mathcal{S}_i} < 0$ or $t=T$, then the final sentences in $\mathrm{\textit{E}}_{t=T}$ are a pseudo evidence-positive set $\mathbb{E}^+$.
To reduce the search space, we empirically set $T = 5$\footnote{Based on observations that 99\% in HotpotQA require less than 6 evidence sentences for reasoning.}.

\begin{figure*}[t]
	\centering
	\includegraphics[width=160mm]{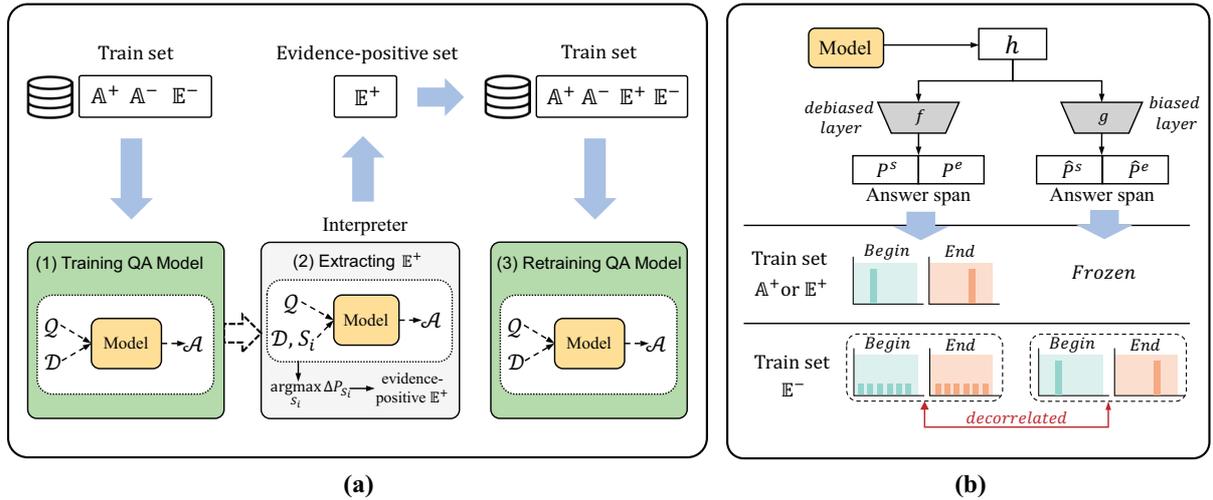}
	\caption{Learning of our proposed approach: (a) Training QA model for evidentiality, extracted by \textit{Interpreter}. (b) Our QA predictor for learning decorrelated features on biased examples.}
	\label{figure2}
\end{figure*}

Briefly, we obtain the labels of answerability and evidentiality, as follows:
\begin{itemize}[leftmargin=0.6cm]
    \item Answer-positive $\mathbb{A}^+$ and negative $\mathbb{A}^-$ set: the former has both answer and evidences, and the latter has neither. 
    \vspace{-1mm}
    \item Evidence-positive $\mathbb{E}^+$ and negative $\mathbb{E}^-$ set: 
    the former is expected to have all the evidences, and the latter has an answer with no evidence.
\end{itemize}

\subsection{Learning Answerability \& Evidentiality}

In this section, our goal is to learn the above labels of answerability and evidentiality.

\subsubsection*{Supervising Answers and Answerability (Base)}

As optimizing QA model is not our focus, we adopt the existing model in \cite{min2019compositional}.
As the architecture of QA modal, we use a powerful transformer-based model -- RoBERTa~\cite{liu2019roberta}, where the input is 
{\fontfamily{qcr}\selectfont[CLS] question [SEP] passage [EOS]}.
The output of the model is as follows:
\begin{equation}
    \label{roberta}
    \begin{aligned}
        & ~~~~~~~~~~~~ h = \text{RoBERTa (Input)} \in \mathbb{R}^{n \times d} \\
        & ~~~~~~~~~~~~~~ O^s = f_1(h), ~~ O^e = f_2(h)\\ 
        & ~~P^s  = softmax(O^s), ~~ P^e  = softmax(O^e) \\
    \end{aligned}
\end{equation}
where $f_1$ and $f_2$ are fully connected layers with the trainable parameters $\in \mathbb{R}^{d}$, $P^s$ and $P^e$ are the the probabilities of start and end positions,
$d$ is the output dimension of the encoder, $n$ is the size of the input sequence.

For answerability, they build a classifier through the hidden state $h_{[0,:]}$ of {\fontfamily{qcr}\selectfont[CLS]} token that represents both $\mathcal{Q}$ and $\mathcal{D}$.
As HotpotQA dataset covers both yes-or-no and span-extraction questions, which we follow the convention of~\cite{asai2019learning} to support both as a multi-class classification problem of predicting the four probabilities:
\begin{equation}
\label{p_class}
\begin{aligned}
    P^{cls}  & = softmax(W_1 h_{[0,:]}) \\  
    & = [p_{span}, ~ p_{yes}, ~ p_{no}, ~ p_{none}] \\
\end{aligned}
\end{equation}
where 
$p_{span}$, $p_{yes}$, $p_{no}$, and $p_{none}$ denote the probabilities of the answer type being span, {\fontfamily{qcr}\selectfont yes}, {\fontfamily{qcr}\selectfont no}, and no answer, respectively, and $W_1 \in \mathbb{R}^{4 \times d}$ is the trainable parameters. 
For training answer span and its class, 
the loss function of example $i$ is the sum of cross entropy losses ($D_{CE}$), as follows:
\begin{equation}
    \begin{aligned}
        ~~ D_{CE}(P_i,\mathcal{A}_i) = -\Big( log(P^s_{s_i}) + log(P^e_{e_i}) \Big)  \\
        D_{CE}(P^{cls}_i, {C}_i) = -log(P^{cls}_{c_i})  ~~~~~~~~\\
        \mathcal{L}_{A}(i)  = D_{CE}(P_i,\mathcal{A}_i) + D_{CE}(P^{cls}_i, {C}_i)  \\
    \end{aligned}
\end{equation}
where $s_i$ and $e_i$ are  the starting and ending position of answer $\mathcal{A}$, respectively, and $c_i$ is the index of the actual class ${C}_i$ in example $i$.

\subsubsection*{Supervising Evidentiality}

As overviewed in Section 1, Base model is reported to take a shortcut, or a direct path between answer $\mathcal{A}$ and question $\mathcal{Q}$, neglecting implicit intermediate paths (evidences).
Specifically, we present the two objectives for unbiased models:
\begin{itemize}[leftmargin=0.5cm]
\item (O1): QA model should not be overconfident on passages with no evidences (\ie, on $\mathbb{E^-}$).
\vspace{-2mm}
\item (O2): QA model should be confident on passages with both answer/evidences (\ie, on $\mathbb{E^+}$)
\end{itemize}

For (O1), as a naive approach, one may consider a regularization term to 
avoid overconfidence on evidence-negative set $\mathbb{E}^-$.
Overconfident answer distribution would be diverged from uniform distribution, such that Kullback–Leibler (KL) divergence $KL(p||q)$,
where $p$ and $q$ are the answer probabilities and the uniform distribution, respectively, is high when overconfident:
\begin{equation}
    \label{regularization}
    \mathcal{R} = \sum_{i ~ \in ~\mathbb{E^-}} D_{KL}( P(\mathcal{A}_i|\mathcal{Q}_i, \mathcal{D}_i) || P_{uniform})
\end{equation}
where $P_{uniform}$ indicates uniform distribution.
This regularization term $\mathcal{R}$ forces the answer probabilities on $\mathbb{E}^-$ to be closer to the uniform one.

However, one reported risk~\cite{utama2020mind,grand2019adversarial} is that
suppressing data with biases has a side-effect of lowering confidence on unbiased data (especially on in-distribution).
Similarly, in our case, regularizing to keep the confidence low for $\mathbb{E}^-$, can cause lowering that for $\mathbb{E}^+$, due to their correlation.
In other words, pursuing (O1) violates (O2), which we observe later in Figure \ref{fig:graphs}.
Our next goal is thus to decorrelate two distributions on $\mathbb{E^+}$ and $\mathbb{E^-}$ to satisfy both (O1) and (O2). 

Figure~\ref{figure2}(b) shows how we feed the hidden states $h$ into two predictors.
Predictor $f$ is for learning the target distribution and predictor $g$
is purposedly trained to be overconfident on evidence-negative set $\mathbb{E}^-$, where this biased answer distribution is denoted as $\hat{P}$.
We regularize target distribution $P$ to diverge from the biased distribution of $\hat{P}$.

Formally, 
the biased answer distributions $\hat{P}$ ($\hat{P}^s$ and $\hat{P}^e$) are as follows:
\begin{equation}
    \label{new_repre}
    \begin{aligned}
        \hat{O^s} = g_1(h), ~~~~ \hat{O^e} = g_2(h) ~~~~~~~~~~~~~~\\
        \hat{P^s}  = softmax(\hat{O^s}), ~~ \hat{P^e}  = softmax(\hat{O^e})\\
    \end{aligned}
\end{equation}
where $g_1$ and $g_2$ are fully connected layers with the trainable parameters $\in \mathbb{R}^{d}$.
Then, we optimize $\hat{P}$ to predict answer $\mathcal{A}$ on evidence-negative set $\mathbb{E^-}$, which makes layer $g$ biased (taking shortcuts), and regularize $f$ by maximizing KL divergence between $P$ and fixed $\hat{P}$.
The regularization term of example $i \in \mathbb{E^-}$ is as follows:
\begin{equation}
    \label{new_repre2}
    \begin{aligned}
        \mathcal{\hat{R}}(i)=  D_{CE}(\hat{P}_i,\mathcal{A}_i) - \lambda D_{KL} (\hat{P}_i || P_i) \\
    \end{aligned}
\end{equation}
where $\lambda$ is a hyper-parameter.
This loss $\mathcal{\hat{R}}$ is optimized on only evidence-negative set $\mathbb{E^-}$.

Lastly, to pursue (O2), we train on $\mathbb{E^+}$, as done on $\mathbb{A^+}$.
However, in initial steps of training, our \textit{Interpreter} is not reliable, since the QA model is not trained enough yet.
We thus train without $\mathbb{E^+}$ for the first $K$ epochs, then extract  $\mathbb{E^+}$ at $K$ epoch and continue to train on all sets, as shown in Figure \ref{figure2}(a).
In the final loss function, we apply different losses as set $\mathbb{E}$ and $\mathbb{A}$:
\begin{equation}
    \label{separate}
    \begin{aligned}
        \mathcal{L}_{total} ~ & = ~  \sum_{i ~ \in ~\mathbb{A^{+,-}}} \mathcal{L}_{A}(i)   +  \sum_{i ~ \in ~\mathbb{E^-}}  \mathcal{\hat{R}}(i) \\
        &  ~~~~  + \sum_{i ~ \in ~ \mathbb{E^+}} u(t-K) \cdot \mathcal{L}_{A}(i) \\
    \end{aligned}
\end{equation}
where the function $u$ is a delayed step function (1 when epoch $t$ is greater than $K$, 0 otherwise).

\subsection{Passage Selection at Inference Time}

For our multi-hop QA task, it requires to find answerable passages with both answer and evidence, from candidate passages.
While we can access the ground-truth of answerability in training set,
we need to identify the answerability of $(\mathcal{Q,D})$ at inference time.
For this, we consider two directions:
(1) Paragraph Pair Selection, which is specific to HotpotQA, and (2) Supervised Evidence Selector trained on pseudo-labels.

For (1), we consider the data characteristic, mentioned in Section 3.1; 
we know one pair of paragraphs is answerable/evidential (when both paragraphs
are positive, or $\mathcal{P}^+$).
Thus, the goal is to identify the answerable pair of paragraphs, from all possible pairs $\mathcal{P}_{ij} = \{(p_i, p_j) : p_i \in \mathcal{P}, p_j \in \mathcal{P}\}$ (denoted as \textbf{paired-paragraph}).
We can let the model select one pair with highest estimated answerability, $1-p_{none}$ in Eq. (\ref{p_class}), and predict answers on the paired passage, which is likely to be evidential.

For (2),
some pipelined approaches~\cite{nie2019revealing,groeneveld2020simple} design an evidence selector, extracting top k sentences from all candidate paragraphs.
While they supervise the model using ground-truth of evidences, we assume there is no such annotation, thus train on pseudo-labels $\mathbb{E^+}$.
We denote this setting as \textbf{selected-evidences}.
For evidence selector, we follow an extracting method in \cite{beltagy2020longformer}, where  
the special token {\fontfamily{qcr}\selectfont[S]} is added at ending position of each sentence, and  $h_{[\fontfamily{qcr}\selectfont S_i]}$ from BERT indicates $i$-th sentence embedding.
Then, a binary classifier $f_{evi}(h_{[\fontfamily{qcr}\selectfont S_i]})$ is trained on the pseudo-labels, 
where $f_{evi}$ is a fully connected layer.
During training, the classifier identifies whether each sentence is evidence-positive (1) or negative (0).
At inference time, we first select top 5 sentences\footnote{Table \ref{statistic} shows the precision and recall of top5 sentences.} on paragraph candidates, and then insert the selected evidences into QA model for testing.

While we discuss how to get the answerable passage above, we can use the passage setting for evaluation.
To show the robustness of our model, we construct a challenge test set by excluding easy examples (\ie, easy to take shortcuts).
To detect such easy examples, we build a set of \textbf{single-paragraph} $\mathcal{P}_i$, that none of it is evidential in HotpotQA,
as the dataset avoids having all evidences in a single paragraph, to discourage single-hop reasoning.
If QA model predicts the correct answer on the (unevidential) single-paragraph, we remove such examples in HotpotQA, and define the remaining set as the challenge set.

\begin{table}[]
\caption{The precision and recall of pseudo evidences from \textit{Interpreter}, compared to the ground truth (GT).}
\label{statistic}
\centering
\scalebox{0.98}{
\begin{tabular}{lccc}
\noalign{\hrule height 1pt} 
 &  \# of sent     & Prec & Recall     \\ \hline
GT evidences  &  2.38   &  100.  &  100.       \\
Answerable $\mathbb{A^+}$  &  6.45   &  36.94  &  100.       \\
$\mathbb{E^+}$ (Train set) &  3.64  &  61.13  &  86.64       \\ 
$\mathbb{E^+}$ (Dev set) &  5.00 &  46.12   & 90.35       \\ 
\noalign{\hrule height 1pt} 
\end{tabular}}
\end{table}

\section{Experiment}

In this section, we formulate our research questions to guide our experiments and describe evaluation results corresponding to each question.

\begin{table*}[]
\caption{The comparison of the proposed models on the original set and challenge set.}
\label{result1}
\centering

\scalebox{0.98}{
\begin{tabular}{llccc}
\noalign{\hrule height 1pt} 
& \multicolumn{1}{c}{\multirow{2}{*}{Model}} & \multicolumn{1}{c}{\multirow{2}{*}{Input at Inference}} & \multicolumn{2}{c}{Question Answering (F1)} \\ \cline{4-5} 
      &     &    &  Original Set      & Challenge Set      \\ 
\noalign{\hrule height 1pt} 

\multicolumn{5}{l}{~ {\textit{without external knowledge}}}  \\ 
\hline
B-\RNum{1}:&  Single-paragraph QA          &   Single-paragraph    &  68.65  &  0.0  \\
B-\RNum{2}: &  Single-paragraph QA          &   Paired-paragraph    &  62.01  & 30.07  \\
O-\RNum{1}:&  Our model               &   Single-paragraph     &  32.61  &  19.81 \\
O-\RNum{2}: &  Our model               &   Paired-paragraph     &  68.08  &  41.69 \\
O-\RNum{3}: & Our model (full)   &   Selected-evidences     &  \textbf{70.21}  &  \textbf{44.57}  \\

\hline
\multicolumn{5}{l}{~{\textit{with external knowledge}}}  \\ 
\hline
C-\RNum{1}: &  \citet{asai2019learning} & ~~ Retrieved-evidences ~~ & 73.30 &  48.54  \\
C-\RNum{2}: &  \citet{asai2019learning} + Ours & Retrieved-evidences & 73.95 & 50.15  \\

\noalign{\hrule height 1pt} 
\end{tabular}}
\end{table*}

\paragraph{Research Questions}
To evaluate the effectiveness of our method, we address the following research questions:
\begin{itemize}[leftmargin=0.6cm]
 \item \textbf{RQ1}: How effective is our proposed method for a multi-hop QA task? 
 \item \textbf{RQ2}: Does our \textit{Interpreter} effectively extract pseudo-evidentiality annotations for training?
 \item \textbf{RQ3}: Does our method  avoid reasoning shortcuts  in unseen data?
\end{itemize}

\paragraph{Implementation}
Our implementation settings for QA model follow RoBERTa (Base version with 12 layers)~\cite{liu2019roberta}.
We use the Adam optimizer with a learning rate of 0.00005 and a batch-size of 8 on RTX titan.
We extract the evidence-positive set after 3 epoch ($K$=3 in Eq. (\ref{separate})) and retrain for 3 epochs.
As a hyper-parameter, we search $\lambda$ among $\{1, 0.1, 0.01\}$, and found the best value ($\lambda$=0.01), based on 5\% hold-out set sampled from the training set.

\paragraph{Metrics}
We report standard F1 score for HotpotQA, to evaluate the overall QA accuracy to find the correct answers. 
For evidence selection, we also report F1 score, Precision, and Recall to evaluate the sentence-level evidence retrieval accuracy.

\subsection{RQ1: QA Effectiveness}

\paragraph{Evaluation Set}
\begin{itemize}[leftmargin=0.4cm]
    \item \textbf{Original Set}: We evaluate our proposed approach on multi-hop reasoning dataset, HotpotQA\footnote{https://hotpotqa.github.io/} \cite{yang2018hotpotqa}.
    HotpotQA contains 112K examples of multi-hop questions and answers.
    For evaluation, we use the HotpotQA dev set (distractor setting) with 7405 examples.
    \item \textbf{Challenge Set}: 
    To validate the robustness, we construct a challenge set where QA model on \textbf{single-paragraph} gets zero F1, while such model achieves 67 F1 in the original set.
    That is, we exclude instances with F1 $>$ 0, where the QA model predicts an answer without right reasoning.
    The exclusion makes sure the baseline obtains zero F1 on the challenge set.
    The number of surviving examples in our challenge set is 1653 (21.5\% of dev set).
\end{itemize}

\begin{table}[]
\caption{The ablation study on our full model.}
\label{ablation}
\centering

\scalebox{0.98}{
\begin{tabular}{lcc}
\noalign{\hrule height 1pt} 
\multicolumn{1}{c}{\multirow{2}{*}{Model}} & \multicolumn{2}{c}{QA (F1)} \\ \cline{2-3} 
                &  Original      & Challenge      \\ \hline
Our model (full)   &  \textbf{70.21}  &  \textbf{44.57}  \\
(A) remove $\mathbb{E^+}$       &  68.51  &  40.78 \\
(B) remove $\mathbb{E^+}$ \& $\mathbb{E^-}$  &  66.42  &  40.75  \\
(C) replace $\mathcal{\hat{R}}$ with $\mathcal{R}$ &  69.64  & 42.54  \\

\noalign{\hrule height 1pt} 
\end{tabular}}
\end{table}

\begin{table*}[]
\caption{The comparison of the proposed models for evidence selection}
\label{result2}
\centering
\scalebox{0.98}{
\begin{tabular}{lccc}
\noalign{\hrule height 1pt} 
\multicolumn{1}{c}{\multirow{2}{*}{Model}} & \multicolumn{3}{c}{Evidence Selection} \\ \cline{2-4} 
                &  F1      & Precision & Recall     \\ \hline
Retrieval-based AIR \cite{yadav2020unsupervised}     &  66.16   &  \textbf{63.06}  &  69.57       \\
Accumulative-based interpreter on our QA model  &  54.05   & 53.56  &  62.38       \\ 
(a) \textit{Interpreter} on Single-paragraph QA  &  56.76   &  57.50  &  63.71       \\
(b) \textit{Interpreter} on our QA model w/ $\mathcal{R}$ &  \textbf{70.30}   &  62.04  &  \textbf{87.10}       \\ 
(c) \textit{Interpreter} on our QA model (full) &  69.35   &  61.09  &  86.59       \\ 
\noalign{\hrule height 1pt} 
\end{tabular}}
\end{table*}

\begin{figure*}[t]
    \centering
    \subfigure[Single-paragraph QA]
    {
        \includegraphics[height=1.1in]{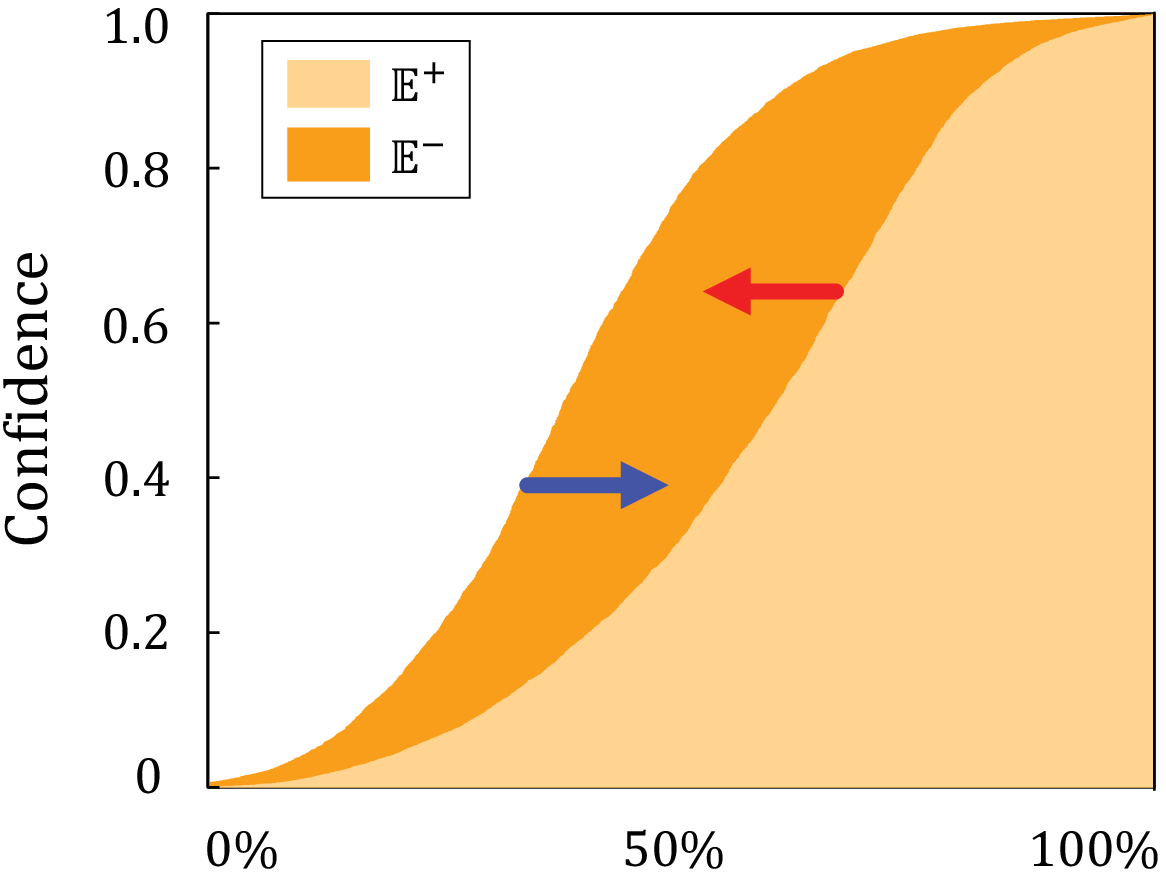}%
        \label{fig:1}
    }
    \subfigure[Ours w/ $\mathcal{R}$]
    {
        \includegraphics[height=1.1in]{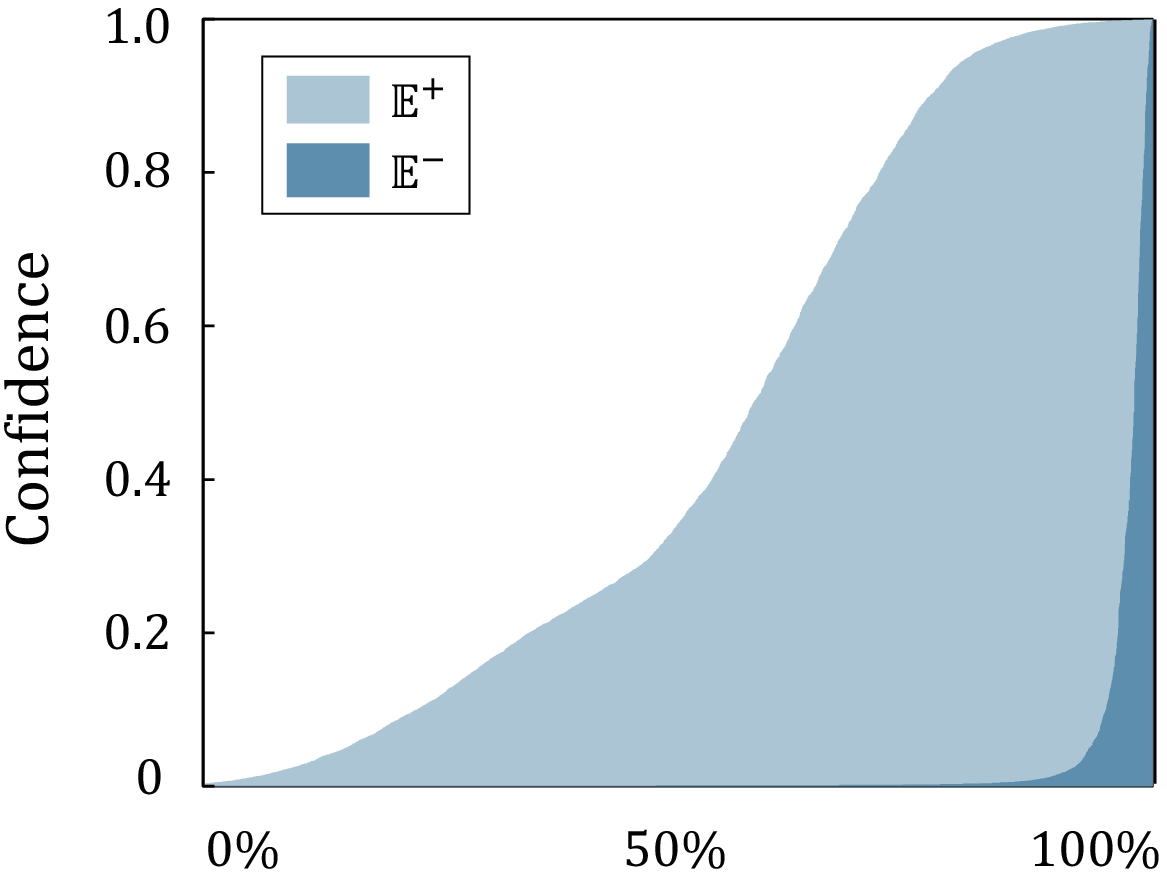}%
        \label{fig:2}
    }
    \subfigure[Ours w/ $\mathcal{\hat{R}}$ (full)]
    {
        \includegraphics[height=1.1in]{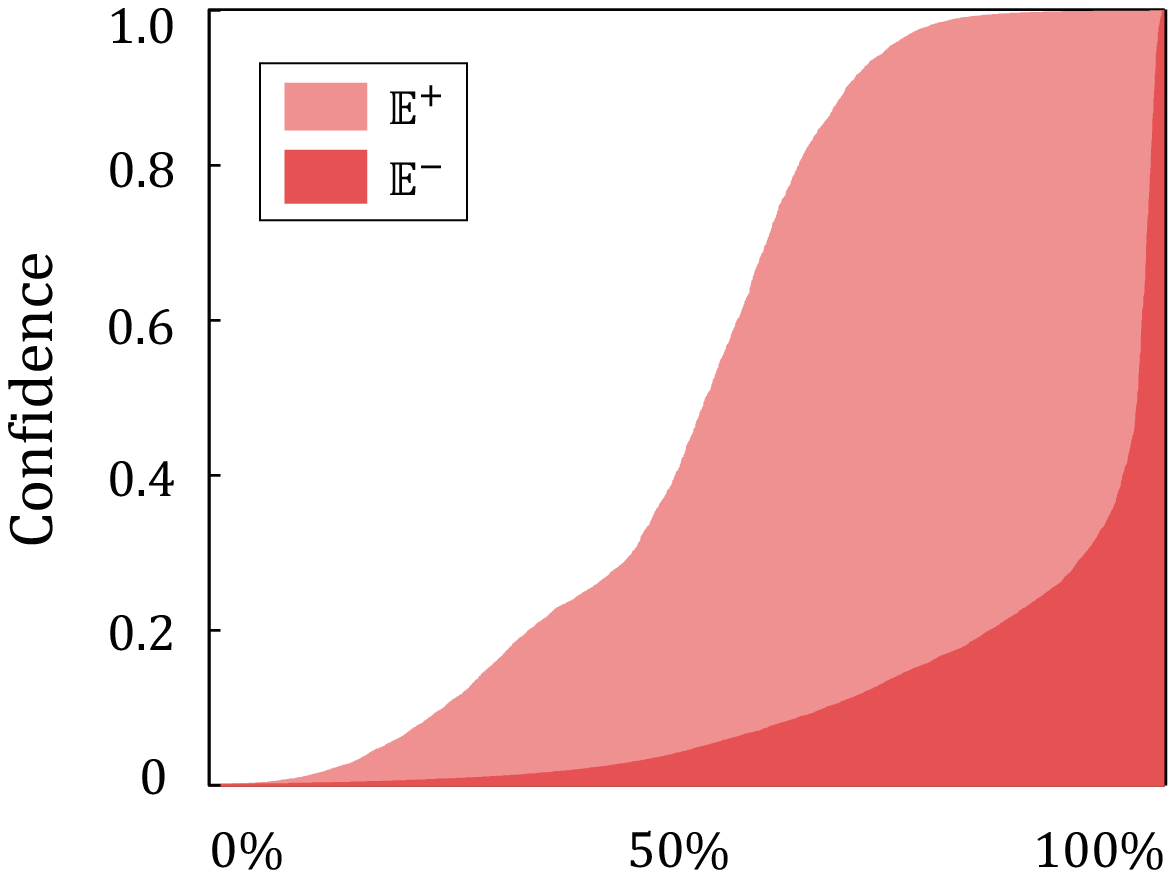}%
        \label{fig:3}
    }
    \subfigure[Three models on $\mathbb{E^+}$]
    {
        \includegraphics[height=1.1in]{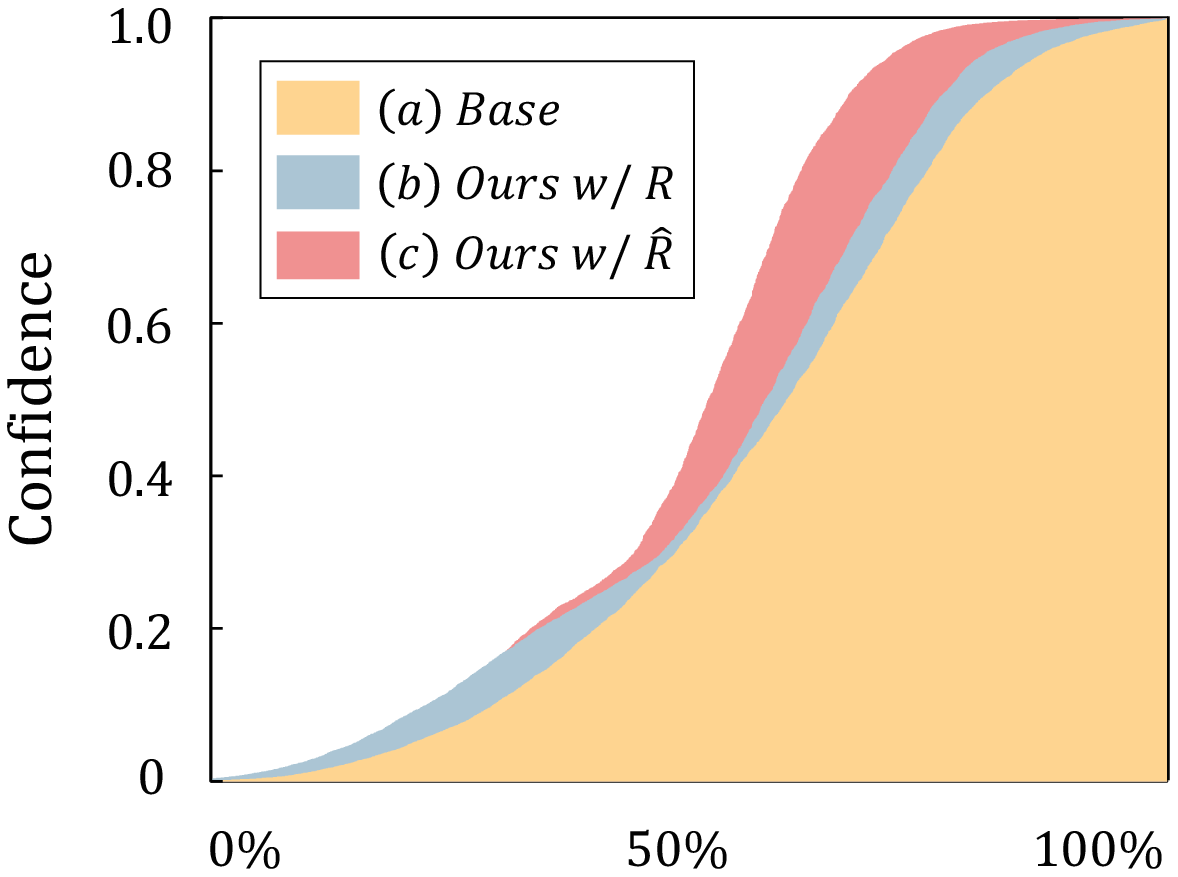}%
        \label{fig:4}
    }
    \caption{\textbf{Confidence Analysis:} Confidence scores of three models in the ascending order, on $\mathbb{E^+}$ (light color) and $\mathbb{E^-}$ (dark colar). (a) Base model trained on single-paragraphs. (b) Our model with $\mathcal{R}$. (c) Our full model with $\mathcal{\hat{R}}$. (d) Comparison of three models on $\mathbb{E^+}$.}
    \label{fig:graphs}
\end{figure*}

\paragraph{Baselines, Our models, and Competitors}

As a baseline, we follow the previous QA model \cite{min2019compositional} trained on single-paragraphs.
We test our model on single-paragraphs, paired-paragraphs and selected evidences settings discussed in Section 3.4.
As a strong competitor, among released models for HotpotQA, we implement a state-of-the-art model~\cite{asai2019learning}\footnote{Highest performing model in the leaderboard of HotpotQA with public code release}, using external knowledge and a graph-based retriever.

\paragraph{Main Results}
This section includes the results of our model for multi-hop reasoning.
As shown in Table \ref{result1}, our full model outperforms baselines on both original and challenge set.

We can further observe that \textbf{\romannumeral 1)} 
when tested on single-paragraphs, where forced to take shortcuts, our model (O-\RNum{1}) is worse than the baseline (B-\RNum{1}), which indicates that B-\RNum{1} learned the shortcuts.
In contrast, O-\RNum{2} outperforms B-\RNum{2} on paired-paragraphs where at least one passage candidate has all the evidences.

\textbf{\romannumeral 2)} When tested on evidences selected by our method (O-\RNum{3}), we can improve F1 scores on both original set and challenge set. This noise filtering effect of evidence selection, by eliminating irrelevant sentences, was consistently observed in a supervised setting~\cite{nie2019revealing,groeneveld2020simple,beltagy2020longformer},  which we could reproduce without annotation.

\textbf{\romannumeral 3)} Combining our method with SOTA (C-\RNum{1})~\cite{asai2019learning} leads to accuracy gains in both sets.
C-\RNum{1} has distinctions of using external knowledge of reasoning paths, to outperform models without such advantages,
but our method can contribute to complementary gains.

\paragraph{Ablation Study}
As shown in Table~\ref{ablation}, we conduct an ablation study of 
O-\RNum{3} in Table \ref{result1}.
In (A), we remove $\mathbb{E^+}$ from \textit{Interpreter}, in training time.
On the QA model without $\mathbb{E^+}$, the performance decreased significantly, suggesting the importance of evidence-positive set.
In (B), we remove evidentaility labels of both $\mathbb{E^+}$ and $\mathbb{E^-}$, and observed that the performance drop is larger compared to other variants.
Through (A) and (B), we show that training our evidentiality labels can increase QA performance.
In (C), we replace $\mathcal{\hat{R}}$ with $\mathcal{R}$, removing layer $g$ to train biased features. 
On the replaced regularization, the performance also decreased, suggesting that training $\mathcal{\hat{R}}$ is effective for a multi-hop QA task.

\subsection{RQ2: Evaluation of Pseudo-Evidentiality Annotation}

In this section, we evaluate the effectiveness of our \textit{Interpreter}, which generates evidences on training set, without supervision.
We compare the pseudo evidences with human-annotation, by sentence-level.
For evaluation, we measure sentence-level F1 score, Precision and Recall, following the evidence selection evaluation in~\cite{yang2018hotpotqa}.

As a baseline, we implement the retrieval-based model, AIR~\cite{yadav2020unsupervised}, which is an unsupervised method as ours.
As shown in Table~\ref{result2}, 
our \textit{Interpreter} on our QA model outperforms the retrieval-based method, in terms of F1 and Recall,
while the baseline (AIR) achieves the highest precision (63.06\%).
We argue recall, aiming at identifying all evidences, is much critical
for multi-hop reasoning, for our goal of avoiding disconnected reasoning,
as long as precision remains higher than precision of answerable $\mathbb{A^+}$ (36.94\%), in Table \ref{statistic}.

As variants of our method, we test our \textit{Interpreter} on various models.
First, when comparing (a) and (c), our full model (c) outperforms the baseline (a) over all metrics.  
The baseline (a) trained on single-paragraphs got biased, thus the evidences generated by the biased model are less accurate.
Second, the variant (b) trained by $\mathcal{R}$ outperforms (c) our full model.
In Eq. (\ref{regularization}), the loss term $\mathcal{R}$ does not train layer $g$ for biased features, unlike $\mathcal{\hat{R}}$ in Eq. (\ref{new_repre2}).
This shows that learning $g$ results in performance degradation for evidence selection, despite performance gain in QA.

\subsection{RQ3: Generalization}

In this section, to show that our model avoids reasoning shortcuts  for unseen data, 
we analyze the confidence distribution of models on the evidence-positive and negative set.
In dev set, we treat the ground truth of evidences as $\mathbb{E^+}$, and a single sentence containing answer as $\mathbb{E^-}$ (each has 7K $\mathcal{Q}$-$\mathcal{D}$ pairs).
On these set, Figure \ref{fig:graphs} shows \textbf{confidence} $P(\mathcal{A|Q,D})$ of three models; (a), (b), and (c) mentioned in Section 4.2.
We sort the confidence scores in ascending order, where y-axis indicates the confidence and x-axis refers to the sorted index.
Thus, the colored area indicates the dominance of confidence distribution.
Ideally, for a debiased model, the area on evidence-positive set should be large, while that on evidence-negative should be small.

Desirably, in Figure \ref{fig:1}, the area under the curve for $\mathbb{E^-}$ should decrease for pursuing (O1), 
moving along \textit{blue} arrow, 
while that of $\mathbb{E^+}$ should increase for (O2), as \textit{red} arrow shows.
In Figure \ref{fig:2}, our model with $\mathcal{R}$ follows \textit{blue} arrow, with a smaller area
under the curve for $\mathbb{E^-}$, while keeping that of $\mathbb{E^+}$ comparable to Figure \ref{fig:1}.
For the comparison, Figure \ref{fig:4} shows all curves on $\mathbb{E^+}$.
In Figure \ref{fig:3}, our full model follows both directions of \textit{blue} and \textit{red} arrows, which indicates that ours satisfied both (O1) and (O2).

\section{Conclusion}
In this paper, we propose a new approach 
to train multi-hop QA models, not to take reasoning shortcuts of guessing right answers without sufficient evidences.
We do not require annotations and generate pseudo-evidentiality instead, by regularizing QA model from being overconfident when evidences are insufficient.
Our experimental results show that our method outperforms baselines on HotpotQA and has the effectiveness to distinguish between evidence-positive and negative set.

\section*{Acknowledgements}


This research was supported by IITP grant funded by the Korea government (MSIT) (No.2017-0-01779, XAI) and ITRC support program funded by the Korea government (MSIT) (IITP-2021-2020-0-01789).





\bibliographystyle{acl_natbib}
\bibliography{acl2021}

\end{document}